\pdfoutput=1

\documentclass[11pt]{article}

\usepackage{ACL2023}

\usepackage{times}
\usepackage{latexsym}

\usepackage{balance}
\usepackage{url}
\usepackage{graphicx}
\usepackage{amsmath,amssymb,amsfonts}
\usepackage{algorithmic}
\usepackage{textcomp}
\usepackage{xcolor}
\usepackage{tikz}
\usetikzlibrary{topaths,calc}
\usepackage[ruled]{algorithm2e}
\usepackage{subfigure}
\usepackage{booktabs}
\usepackage{balance}
\usepackage{colortbl}
\usepackage{enumitem}
\usepackage{multirow}
\usepackage[normalem]{ulem}

\usepackage[T1]{fontenc}

\usepackage[utf8]{inputenc}

\usepackage{microtype}

\usepackage{inconsolata}

\newcommand{\revise}[1]{#1}
\newcommand{\revisenew}[1]{#1}

\title{Can Language Models Be Specific? How?}

\author{Jie Huang$^{1}$ $\quad$ Kevin Chen-Chuan Chang$^{1}$ $\quad$ Jinjun Xiong$^{2}$ $\quad$ Wen-mei Hwu$^{1,3}$ \\
 $^1$University of Illinois at Urbana-Champaign, USA \\
 $^2$University at Buffalo, USA \\
 $^3$NVIDIA, USA \\
 \texttt{\{jeffhj, kcchang, w-hwu\}@illinois.edu} \\
 \texttt{jinjun@buffalo.edu}
}

\begin{document}
\maketitle
\begin{abstract}
``\textit{He is a person}'', ``\textit{Paris is located on the earth}''. Both statements are correct but meaningless -- due to lack of specificity.
In this paper, we propose to measure how specific the language of pre-trained language models (PLMs) is. To achieve this, we introduce a novel approach to build a benchmark for specificity testing by forming masked token prediction tasks with prompts. For instance, given ``Toronto is located in [MASK].'', we want to test whether a more specific answer will be better filled in by PLMs, e.g., \textit{
Ontario} instead of \textit{Canada}.

From our evaluations, we show that existing PLMs have only a slight preference for more specific answers. We identify underlying factors affecting the specificity and design two prompt-based methods to improve the specificity. Results show that the specificity of the models can be improved by the proposed methods without additional training. We hope this work can bring to awareness the notion of specificity of language models and encourage the research community to further explore this important but understudied problem.\footnote{Code and data are available at \url{https://github.com/jeffhj/S-TEST}.}
\end{abstract}

\section{Introduction}

Pre-trained language models (PLMs) such as BERT \cite{devlin2019bert} and GPT-2/3 \cite{radford2019language,NEURIPS2020_1457c0d6} have achieved quite impressive results in various natural language processing tasks.
Recent works show that the parameters of these models contain significant amounts of knowledge \cite{petroni2019language,roberts2020much,jiang2020x,jiang2020can,wang2020language}, and knowledge stored in PLMs can be extracted by predicting the mask token(s) using prompts. 
For instance, given prompt ``J. K. Rowling was born in [MASK].'', PLMs can predict the birthplace of Rowling based on its knowledge.

\begin{figure}[tp!]
\centerline{\includegraphics[width=0.8\linewidth]{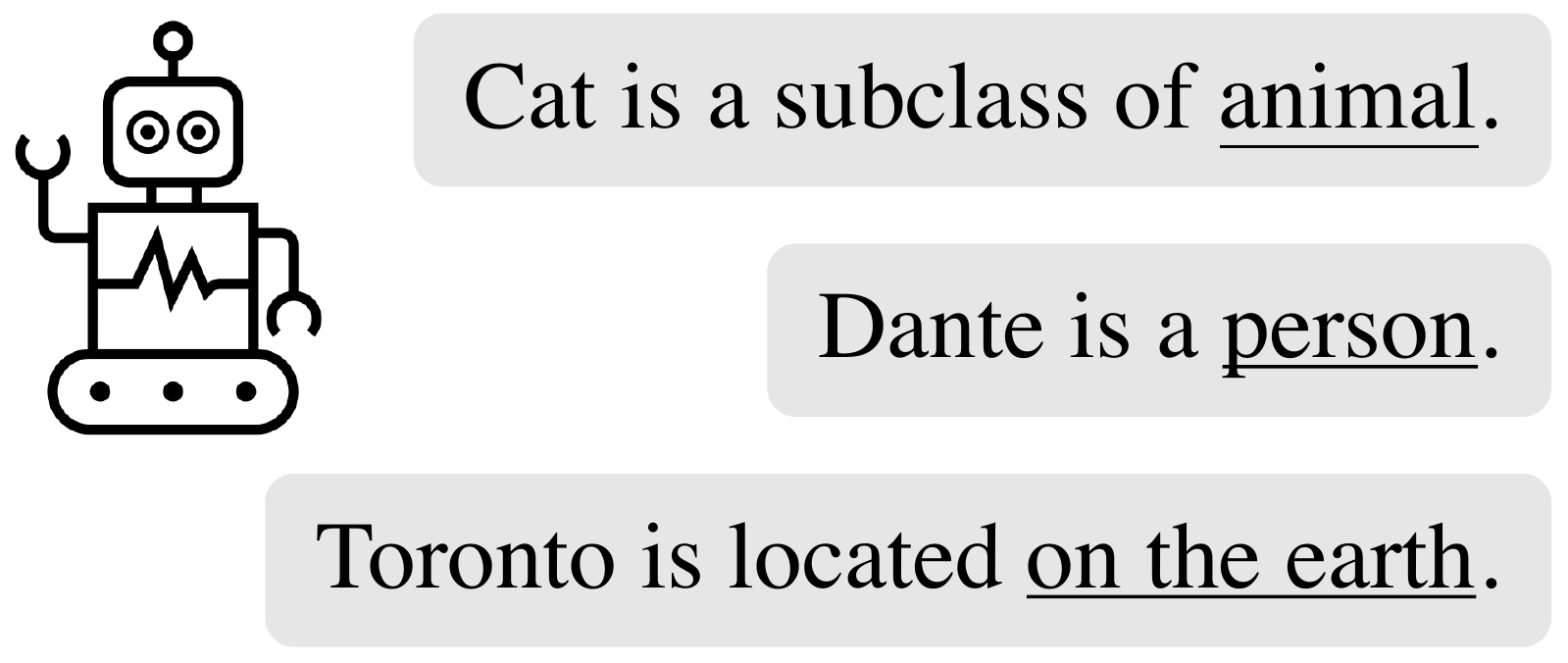}}
\vspace{-1mm}
\caption{Examples of language modeling that lack specificity. More specific descriptions could be: \uline{feline}, \uline{poet}, and \uline{in Ontario}, respectively.}
\label{fig:intro}
\vspace{-3.5mm}
\end{figure}

However, there may exist multiple answers for a query, while not all answers are equally specific. \revise{In many situations, we desire a specific answer.}
For the example above, the masked token can be replaced by \textit{Yate} (a town), \textit{Gloucestershire} (a county), or \textit{England} (a country). 
\revise{To acquire the maximum knowledge (in this example, the town, the county, and the country where Rowling was born), 
we may prefer the model to fill in \textit{Yate} since \textit{Gloucestershire} and \textit{England} can be further predicted using prompts, e.g., ``Yate is located in [MASK].''}
\revise{This means, if the prediction is more specific, we can retrieve more fine-grained information from language models, and further acquire more information.}
Besides, sometimes, the less specific answer is not useful. For instance, it is well known that \textit{Chicago} is located in \textit{the USA}, users will not get additional information if the model only predicts \textit{Chicago} is located in \textit{the USA} instead of \textit{Illinois}. More examples are shown in Figure \ref{fig:intro}.
To make an analogy:
\revisenew{A good speaker not only needs to be correct, but also has the ability to be specific when desired. The same is true for language models.}

Although there are works on measuring how much knowledge is stored in PLMs or improving the \textit{correctness} of the predictions \cite{petroni2019language,roberts2020much,jiang2020can}, few attempted to measure or improve the \textit{specificity} of predictions made by PLMs. 
Noteworthy exceptions include the work by \citet{adiwardana2020towards,thoppilan2022lamda}, who evaluated the specificity of conversational language models. In their research, specificity was defined and measured within a conversational context -- for instance, the response ``Me too. I love Eurovision songs'' is deemed more specific than simply ``Me too'' to the statement ``I love Eurovision''.
Understanding how specific the language of PLMs is can help us better understand the behavior of language models and facilitate downstream applications such as question answering, text generation, and information extraction \cite{liu2021pre,khashabi2020unifiedqa,NEURIPS2020_1457c0d6,wang2020language}, e.g., making the generated answers/sentences or extracted information more specific or fine-grained.

Therefore, we propose to build a benchmark to measure the specificity of the language of PLMs. 
For reducing human effort and easier to further expand the dataset (e.g., to specific domains),
we introduce a novel way to construct test data automatically based on transitive relations in Wikidata \cite{vrandevcic2014wikidata}.
Specifically, we extract reasoning paths from Wikidata, e.g., (\text{J. K. Rowling}, \textit{birthplace}, \text{Yate}, \textit{location}, \text{Gloucestershire}, \textit{location}, \textit{England}).
Based on the average distance of each object to the subject and the property of transitive relations, we form masked-token-prediction based probing tasks to measure the specificity, e.g., whether the masked token in ``J. K. Rowling was born in [MASK].'' is better filled by \textit{Yate} than \textit{England} by PLMs.
The resulting benchmark dataset contains more than \texttt{20,000} probes covering queries from 5 different categories.
The quality of the benchmark is high, where the judgment \revise{on which answer is more specific} is $\sim97\%$ consistent with humans.

We provide in-depth analyses on model specificity and study two factors that affect the specificity with our benchmark. 
As shown by our evaluations in Section \ref{sec:analysis}, existing PLMs, e.g., BERT and GPT-2, similarly have only a slight preference for more specific answers (in only about $60\%$ of cases where a more specific answer is preferred). 
We also show that, in general, PLMs prefer less specific answers without subjects given, and they only have a weak ability to differentiate coarse-grained/fine-grained objects by measuring their similarities to subjects.
The results indicate that specificity was neglected by existing research on language models.
\revisenew{How to improve and control it is undoubtedly an interesting and valuable problem.}

Based on our observations and analyses, we propose two techniques to improve the specificity of the predictions by modifying the prompts without additional training: 
\textbf{\textit{Few-shot Prompting}}, where demonstrations with more specific answers are provided to guide the models to produce more specific answers; and \textbf{\textit{Cascade Prompting}}, where \textit{which clauses} are added as suffixes to bias the predictions to be more specific. Results show that Few-shot Prompting can improve the specificity for unidirectional language models like GPT-2 well, while Cascade Prompting works well for bidirectional language models such as BERT.

The main contributions of our work are summarized as follows:
\begin{itemize}[nolistsep]
\item We propose a novel automatic approach to build a benchmark for specificity testing based on the property of transitive relations.
\item \revisenew{We analyze the specificity of several existing PLMs and study two factors that affect the specificity.}
\item We propose two methods to improve the specificity by modifying the prompts without additional training.
\item We provide in-depth analyses and discussions, suggesting further works to explore and further improve the specificity.
\end{itemize}

\section{Background and Related Work}

{\flushleft \textbf{Pre-Trained Language Models}:}
Pre-trained language models (PLMs) are language models pre-trained on large corpora.
In this paper, we will cover two types of pre-trained language models: unidirectional language models, such as GPT-2 \cite{radford2019language}, where the prediction of the current token is only based on previous tokens; and bidirectional language models, such as BERT \cite{devlin2019bert} and RoBERTa \cite{liu2019roberta}, where both left and right contexts are utilized to predict the current token.

{\flushleft \textbf{Knowledge Retrieval from LMs and Prompting}:}
Previous works have worked on extracting factual knowledge from PLMs without incorporating external knowledge, which is usually achieved by creating prompts and letting PLMs predict the masked token(s) \cite{petroni2019language,bouraoui2020inducing,jiang2020x,jiang2020can,wang2020language}.
They demonstrated that PLMs contain a significant amount of knowledge. By creating appropriate prompts with some additional training, such methods can even achieve performance comparable to SOTA for some specific tasks \cite{shin2020eliciting,liu2021gpt}.
Our work is inspired by these works; but different from these works, where the focus is to measure or improve the \textit{correctness} of the predictions, our work focuses on measuring and improving the \textit{specificity} of the predictions.

\section{S-TEST: Specificity Testing}

In this section, we introduce our specificity testing (S-TEST) task, describe the creation process of the dataset, and design the metric to measure the specificity of predictions made by PLMs.

\begin{table*}[h]
\setlength\tabcolsep{4pt} %
\small
\begin{center}
\begin{tabular}{l|l|r|l|c|c}
\toprule
 \textbf{ID} & \textbf{Relation} & \textbf{Number} & \textbf{Prompt} & \textbf{Answer 1} & \textbf{Answer 2}  \\
\midrule 
P19 & birthplace & 5,000 & John G. Bennett was born in [MASK]. & London & England \\
\hline
P106 & occupation & 5,000 & Jenny Burton is a [MASK] by profession. & singer & musician \\
\hline
P131 & location & 5,000 & Carey River is located in [MASK]. & Victoria & Australia \\
\hline
P279 & subclass-of & 5,000 & Tracking ship is a subclass of [MASK]. & vessel & vehicle \\
\hline
P361 & part-of & 628 & Hard palate is part of [MASK]. & mouth & head \\
\bottomrule
\end{tabular}
\end{center}
\vspace{-1mm}
\caption{Statistics and examples of the S-TEST benchmark, where we use the same templates in \citet{petroni2019language} to create prompts. \textit{Answer 1} is more specific than \textit{Answer 2}.}
\label{table:examples}
\vspace{-4mm}
\end{table*}

\subsection{Task Formulation}

Specificity is a semantic feature of language to describe things specifically in a given context.
In this work, we focus on measuring the specificity of the predictions produced by pre-trained language models for entity relations. 
Formally, if $(x, r, y) \land (y, r, z)$ implies $(x, r, z)$, then $y$ is considered as a more fine-grained object of $x$ than entity $z$ under relation $r$, and $y$ is more specific than $z$.
For instance, to extract the answer (object) for relation (Toronto, \textit{location}, \textbf{X}),
we convert the query to a masked token prediction task using prompts, e.g., ``Toronto is located in [MASK].'' and let PLMs predict the masked token.
The answer here can be a coarse-grained one, e.g., \textit{Canada}, or a fine-grained one, e.g., \textit{Ontario}. 
\revisenew{The model is considered to be more specific if it tends to fill in \textit{Ontario} instead of \textit{Canada}.}
More general scenarios are discussed in Section \ref{sec:discussion} as future work.

\subsection{Test Data Construction}

We build a benchmark dataset for measuring the specificity based on Wikidata \cite{vrandevcic2014wikidata}, which is a knowledge base containing a large number of entities and relations. 
Specifically,
we utilize transitive relations\footnote{\url{https://www.wikidata.org/wiki/Wikidata:List_of_properties/transitive_relation}} in Wikidata to create the test set automatically. Transitive relations are binary relations with properties such that $(x,\textit{r},y)$ and $(y,\textit{r},z)$ implies $(x,\textit{r},z)$, where entity $y$ can be considered as a more fine-grained object of $x$ than entity $z$ under relation $r$. 

For instance, relation \textit{P131} is a transitive relation, whose label is ``located in the administrative territorial entity''. From Wikidata, we can extract facts (\text{Toronto}, \textit{P131}, \text{Ontario}) and (\text{Ontario}, \textit{P131}, \text{Canada}), which furthermore forms a reasoning path (\text{Toronto}, \textit{P131}, \text{Ontario}, \textit{P131}, \text{Canada}). And \textit{Ontario} is considered more fine-grained (specific) than \textit{Canada} in terms of relation \textit{P131} because its distance to \textit{Toronto} is shorter than \textit{Canada} in the reasoning path.
Based on this, for a transitive relation, we collect reasoning paths with length $\leq5$ for each subject and calculate the average distance of each object to the subject. \revise{E.g., if there are two reasoning paths connecting the subject and object, with lengths 2 and 3, the average distance is 2.5.}
In this way, we can construct pairs with coarse-grained/fine-grained objects for each subject, e.g., (\text{Toronto}, \text{Ontario}) and (\text{Toronto}, \text{Canada}) for \textit{Toronto} in terms of relation \textit{P131} (or a triplet denoted as (\text{Toronto}, \text{Ontario}, \text{Canada})).
The constructed pairs can be used to test the specificity with prompt: ``Toronto is located in [MASK].''

We also combine different relations to form tasks. For instance, for relation \textit{P19}, whose label is ``place of birth'', we combine it with \textit{P131} and further form a mask token prediction task, such as ``[X] was born in [MASK].'' An example reasoning path containing coarse-grained/fine-grained objects is (\text{John G. Bennett}, \textit{P19}, \text{London}, \textit{P131}, \text{England}), corresponding to pairs (John G. Bennett, London) and (John G. Bennett, England).

Considering the representativeness and comprehensiveness, we select 5 relations (Table \ref{table:examples}) and randomly sample up to 5,000 pairs for each relation, with the difference of average distance of the objects to the subject being greater than or equal to $1$ (to filter out entity pairs whose specificity is difficult to differentiate). 
Similar to \citet{petroni2019language}, we only choose single-token objects as the prediction targets, since multi-token generation is still an area that needs further exploration, and the multi-token decoding process will introduce many tunable parameters that obscure the performance \cite{welleck2019non,jiang2020x}.
Statistics and examples of the resulting benchmark dataset are shown in Table \ref{table:examples}. 

\subsection{Metric}

If a model tends to be more specific, 
it should have higher confidence that the more specific answer is correct. For instance, given ``Toronto is located in [MASK].'', the model should assign a higher probability for \textit{Ontario} than \textit{Toronto}.
Therefore, we can measure the specificity by calculating how much times the probability of the fine-grained answer is higher than that of the coarse-grained answer:
\begin{equation*}
\label{eq:metric}
    p_r = \frac{1}{|\mathcal{T}_r|} \sum_{(x, y_1, y_2) \in \mathcal{T}_r} \mathbf{1}[c(y_1|x,r)>c(y_2|x,r)],
\end{equation*}
where $\mathcal{T}_r$ is the set of test examples for relation $r$. $y_1$ is the fine-grained object and $y_2$ is the coarse-grained object. $c(y|x,r)$ is the probability of the model with $y$ as the prediction of the masked token, and $x$ refers to the subject. 
$p_r$ ranges from $0$ to $1$, and $0.5$ means the model does not have a preference in terms of specificity.
The metric is similar to the one used in \citet{marvin2018targeted}, which compares the probability of a pair of words for creating a grammatical sentence, e..g, \textit{The author \uline{laughs}} (grammatical) vs \textit{The author \uline{laugh}} (ungrammatical).

\section{Analysis}

\label{sec:analysis}

\begin{table*}[h]
\begin{center}
\setlength\tabcolsep{4pt} %
\small
\begin{tabular}{l|ccccc|c}
\toprule
& \textbf{birthplace} & \textbf{occupation} & \textbf{location} & \textbf{subclass-of} & \textbf{part-of} & \textit{\textbf{Average}} \\
\midrule 
GPT-2 & 59.72 & 57.28 & 48.25 & 57.98 & 60.86 & 56.82 \\
\hline
BERT-Base & 60.68 & 70.46 & 49.09 & 67.64 & 67.41 & 63.06 \\
\hline
BERT-Large & 56.52 & 71.76 & 42.36 & 77.25 & 66.77 & 62.93 \\
\hline
RoBERTa-Base & 54.48 & 61.80 & 49.99 & 61.59 & 59.11 & 57.39 \\
\hline
RoBERTa-Large & 42.16 & 71.44 & 43.28 & 80.63 & 59.27 & 59.36 \\
\bottomrule
\end{tabular}
\end{center}
\vspace{-1mm}
\caption{\revise{Results of specificity testing with $p_r (\%)$.}}
\label{table:p_g_pre}
\end{table*}

\begin{table*}[h]
\begin{center}
\setlength\tabcolsep{4pt} %
\small
\begin{tabular}{l|ccccc|c}
\toprule
& \textbf{birthplace} & \textbf{occupation} & \textbf{location} & \textbf{subclass-of} & \textbf{part-of} & \textit{\textbf{Average}} \\
\midrule 
\cellcolor{white}\textit{Freq} & \textit{85.87} & \textit{52.86} & \textit{95.11} & \textit{51.12} & \textit{49.68} & \textit{66.93} \\
\hline
\cellcolor{white}\textit{Human} & \textit{98.75} & \textit{92.50} & \textit{100.00} & \textit{96.25} & \textit{97.75} & \textit{97.05} \\
\bottomrule
\end{tabular}
\end{center}
\vspace{-1mm}
\caption{\revise{Results of \textit{Freq} and \textit{Human}.}}
\vspace{-3mm}
\label{table:freq_human}
\end{table*}

In this section, we first analyze the results of S-TEST and then identify and study two underlying factors that affect the specificity of predictions produced by pre-trained language models.

\subsection{Experimental Setup}

\label{sec:setup}

We test on the following pre-trained case-sensitive language models: GPT-2, BERT-Base, BERT-Large, RoBERTa-Base, and RoBERTa-Large. 
For a fair comparison, following \cite{petroni2019language}, we use the intersection of the vocabularies of all the models as the unified vocabulary for prediction ($\sim$18k case-sensitive tokens).
\revise{Since fine-grained answers may be used less frequently in the corpus (e.g., \textit{Yate} is much less frequent than \textit{England}), we also design a simple method by filling the masked tokens with less frequent answers (\textit{Freq}).\footnote{The frequency is calculated with Wikipedia dump \url{https://dumps.wikimedia.org/enwiki/}.} }

\revise{To verify the quality of the dataset,}
we randomly sampled 400 examples (80 for each relation) and asked human annotators to fill in the masked token with both the coarse-grained and fine-grained answers provided \revise{(the order of answers in each pair is randomly shuffled)}. 
\revise{For example, we give annotators both query ``Toronto is located in [MASK].'' and answer pair (Ontario, Toronto) and ask them to select the more specific one. Humans can make judgments based on their own knowledge or relevant information about the entities on the Web.}

\subsection{Results of S-TEST}

\begin{table*}[h]
\begin{center}
\small
\setlength\tabcolsep{4pt} %
\begin{tabular}{l|c|c|c|c|c}
\toprule
& GPT-2 & BERT-Base & BERT-Large & RoBERTa-Base & RoBERTa-Large \\
\midrule 
\textit{Acc@$10$} & 23.87 & 42.65 & 46.80 & 30.79 & 31.81 \\
\bottomrule
\end{tabular}
\end{center}
\vspace{-1mm}
\caption{\revise{The correctness of the predictions measured with \textit{Acc@$10$} $(\%)$.}}
\label{table:correctness}
\end{table*}

Table \ref{table:p_g_pre} reports the results of specificity testing. 
\revisenew{We observe that existing pre-trained language models have only a slight preference for more specific answers, where the probability that more specific answers are preferred by them is around $60\%$.}
This is reasonable since the training of PLMs does not introduce any constraint/bias in terms of specificity.

In Table~\ref{table:freq_human}, the \textit{Freq} method performs quite well on relation \textit{birthplace} and \textit{location} whose answers are both locations, which indicates low frequency may hinder outputting more specific concepts.
However, for other relations, the results are close to random guess.
We also observe that the results of ``\textit{human}'' is high, which demonstrates that the quality of the dataset is high.

\revise{To investigate the correctness of the predictions as in \citet{petroni2019language}, we also calculate \textit{Acc@$10$} (the value is 1 if the coarse/fine-grained answer is ranked among the top 10 results, which are selected from $\sim$18k tokens, and 0 otherwise) among all relations in Table \ref{table:correctness}. We draw a conclusion similar to \citet{petroni2019language} that PLMs have a good ability to recover factual knowledge.\footnote{\revise{The results can be further improved by using techniques such as in \cite{jiang2020can} or applying more advanced language models such as GPT-3 \cite{NEURIPS2020_1457c0d6} -- not the focus of this paper.}}}

Another interesting finding is that for a single relation, the specificity of different models is highly correlated. For instance, for relation \textit{location}, the specificity measured by $p_r$ of all models is slightly lower than $50\%$, while for relation \textit{part-of}, the specificity of all models is around $60\%$. The average pairwise Pearson correlation coefficient among all relations (calculated between different rows) is $0.803$. We think this is because these PLMs are trained on large general corpora; therefore, their knowledge overlaps to a large extent, as is the preference on the specificity of predictions. 

\begin{table*}[h]
\begin{center}
\small
\setlength\tabcolsep{4pt} %
\begin{tabular}{l|l|ccccc|c}
\toprule
\multicolumn{2}{c|}{} & \textbf{birthplace} & \textbf{occupation} & \textbf{location} & \textbf{subclass-of} & \textbf{part-of} & \textit{\textbf{Average}} \\
\midrule 
\multirow{2}{*}{GPT-2} & Naturalness & 46.42 & 50.86 & 10.94 & 60.06 & 51.12 & 43.88 \\
& Relatedness & 68.51 & 78.50 & 82.84 & 40.00 & 50.16 & 64.00 \\
\hline
\multirow{2}{*}{BERT-Base} & Naturalness & 64.81 & 75.04 & 4.99 & 47.96 & 50.80 & 48.72 \\
& Relatedness & 74.89 & 51.96 & 76.43 & 71.67 & 58.79 & \textbf{66.75} \\
\hline
\multirow{2}{*}{BERT-Large} & Naturalness & 66.35 & 79.22 & 10.03 & 48.92 & 47.60 & \textbf{50.42} \\
& Relatedness & 54.46 & 49.16 & 56.22 & 72.96 & 65.50 & 59.66 \\
\hline
\multirow{2}{*}{RoBERTa-Base} & Naturalness & 44.80 & 61.12 & 23.27 & 42.06 & 36.90 & 41.63 \\
& Relatedness & 68.73 & 58.50 & 65.73 & 39.51 & 56.87 & 57.87 \\
\hline
\multirow{2}{*}{RoBERTa-Large} & Naturalness & 31.37 & 66.24 & 3.67 & 43.64 & 41.69 & 37.32 \\
& Relatedness & 47.82 & 41.32 & 34.89 & 55.17 & 64.22 & 48.68 \\
\bottomrule
\end{tabular}
\end{center}
\vspace{-2mm}
\caption{Relatedness and naturalness measured with $p_r (\%)$.}
\label{table:factors}
\vspace{-5mm}
\end{table*}

\subsection{Factors Affecting Specificity}

Some types of questions may be answered specifically naturally. For instance, when discussing anyone's occupation, people may be inclined to use a more specific description; but for the location of a place, people may not be so. 
In addition, specific answers may be easier to relate to the entities in the query than the coarse-grained ones since their connections may be more close, e.g., $similarity(\text{Toronto},\text{Ontario})>similarity(\text{Toronto},\text{Canada})$.
\revise{In this case, the models should tend to select more specific answers.}
Based on the above analysis, the specificity of the predictions mainly depends on question types (e.g., relations) and entities in the query (e.g., subjects), which is also indicated by the metric for measuring specificity, i.e., $c(y|x,r)$. To investigate the effect of each component, 
we split the query, e.g., ``\textcolor{orange}{Toronto} \textcolor{blue}{is located in} [MASK].'', into two parts: the relations, e.g., \textcolor{blue}{\textit{is located in}}, and the subjects, e.g, \textcolor{orange}{\textit{Toronto}}, corresponding to \textit{naturalness} and \textit{relatedness} respectively.

{\flushleft \textbf{\textit{Naturalness}}:} For some questions, they may be answered more specifically naturally than others by PLMs. For instance, for questions about the place of birth, if in the corpora, the birthplace is usually described more specifically, 
e.g., \textit{... was born in Honolulu, Hawaii}, 
PLMs will also describe the birthplace more specifically. 
This is intuitive since PLMs are trained on large corpora based on tasks like masked language modeling; therefore, it will produce more fine-grained predictions conditioned with contexts that are more likely to associate with specific answers. 

To measure how natural a type of questions will be answered more specifically by PLMs, we mask the subject in each prompt, e.g., ``[MASK] was born in [MASK].'', and let PLMs predict the second masked token. 
We get the probability of each token in the vocabulary, i.e., $c(y|\cdot,r)$, and use our metric and dataset to measure the naturalness, e.g., how natural birthplace will be described more specifically in general.

{\flushleft \textbf{\textit{Relatedness}}:} 
Considering the following situation:
\revise{the model can predict that both \textit{A} and \textit{B} are likely to be the correct answers, and judges \textit{A} is more related to the the subject than \textit{B} in general. Intuitively, it will prefer answer \textit{A}.}

Therefore, another factor that affects the specificity of predictions made by PLMs is \textit{relatedness}, i.e., to what extent are the fine-grained objects more related to the corresponding subjects than the coarse-grained ones considered by PLMs. (More generally, this is the ability of PLMs to identify more related entities).

We measure relatedness with phrase embeddings from PLMs. Following \citet{yu2020assessing,wang2021phrase}, we use the mean-pooled representations over the ﬁnal-layer outputs from PLMs as phrase embeddings, and calculate the cosine similarities between the subject and the corresponding objects. If the cosine similarity between the subject and the fine-grained object is higher than that between the subject and the coarse-grained object, we think PLMs consider the fine-grained one is more related to the subject. According to this, we can use our metric and dataset to measure the relatedness, with confidence, i.e., $c(y|x,\cdot)$, based on cosine similarity between $x$ and $y$.

\paragraph{\textbf{Findings.}} 
In Table \ref{table:factors}, we report the \textit{naturalness} and \textit{relatedness} with $p_r$ as the metric. We find that, 
1) the highest average naturalness and relatedness are achieved by BERT-Large and BERT-Base, respectively, corresponding to the highest average specificity; 
2) in many cases, naturalness is lower than $0.5$, which indicates that, without the subjects provided, PLMs are more likely to provide coarse-grained answers, \revise{we think this is because a single coarse-grained entity encompasses the probability mass of many fine-grained entities}; 
3) relatedness is usually higher than $0.5$, which means PLMs have a certain ability to distinguish fine-grained/coarse-grained answers based on semantic similarities between entities.
But the ability is weak since the average scores are just around $60 \%$.

\vspace{-1mm}
\section{Can Language Models Be \textit{MORE} Specific?}
\vspace{-1mm}
\label{sec:method}

\revisenew{From the previous sections, we observe that existing pre-trained language models do not have much preference for more specific answers in a vanilla setting.}
We also observe that PLMs achieve \textit{naturalness} lower than $0.5$, i.e., naturally, PLMs tend to fill in coarse-grained answers with respect to certain types of questions, and \textit{relatedness} around $0.6$, i.e., PLMs only have a weak ability to distinguish more related entities.
Naturalness depends on both the parameters of PLMs and prompts while relatedness only depends on the parameters of PLMs. 
Since it is expensive to change the parameters of PLMs (both time and space), to improve the specificity, we focus on improving the naturalness by modifying the prompts.

Intuitively, to get more specific answers, a practical approach is to ask more specific questions. For instance, to know where Toronto is located more specifically, we may change the prompt ``Toronto is located in [MASK].'' to ``Toronto is located in the province of [MASK].'' However, to achieve this, humans are required to have additional knowledge, e.g., Toronto is a city, and in Canada, the administrative unit larger than city is province rather than state. 
Besides, designing such manually crafted prompts can also be time-consuming and laborious if there are a large number of queries. Furthermore, some questions may be difficult to ask more specifically. For instance, for question ``Hard palate is part of [MASK].'', it is not easy to come up with a more specific query.

Based on the above considerations, we propose two novel and simple techniques to improve the specificity of the predictions. The proposed methods can apply to different models on various types of queries while no additional training is required.

\begin{table*}[h]
\begin{center}
\small
\begin{tabular}{l|l}
\toprule
 \textbf{Relation} & \textbf{Prompt} \\
\midrule 
birthplace & \textcolor{violet}{John G. Bennett was born in [MASK],} \textcolor{brown}{which is located in [MASK].} \\
\hline
occupation & \textcolor{violet}{Jenny Burton is a [MASK] by profession,} \textcolor{brown}{which belongs to [MASK].} \\
\hline
location & \textcolor{violet}{Carey River is located in [MASK],} \textcolor{brown}{which is located in [MASK].} \\
\hline
subclass-of & \textcolor{violet}{Tracking ship is a subclass of [MASK],} \textcolor{brown}{which is a subclass of [MASK].} \\
\hline
part-of & \textcolor{violet}{Hard palate is part of [MASK],} \textcolor{brown}{which is part of [MASK].} \\
\bottomrule
\end{tabular}
\end{center}
\vspace{-2mm}
\caption{Example prompts for Cascade Prompting.}
\vspace{-2mm}
\label{table:cascade_prompts}
\end{table*}

\begin{table*}[h]
\begin{center}
\small
\setlength\tabcolsep{4pt} %
\begin{tabular}{l|ccccc|c}
\toprule
& \textbf{birthplace} & \textbf{occupation} & \textbf{location} & \textbf{subclass-of} & \textbf{part-of} & \textit{\textbf{Average}} \\
\midrule 
GPT-2 (VP) & 59.72 & 57.28 & 48.25 & 57.98 & \textbf{60.86} & 56.82 \\
GPT-2 (FP) & \textbf{81.01} & \textbf{71.66} & \textbf{50.33} & \textbf{64.15} & 57.67 & \textbf{64.96} \\
GPT-2 (CP)* & 59.72 & 57.28 & 48.25 & 57.98 & \textbf{60.86} & 56.82 \\
\hline
BERT-Base (VP) & 60.68 & 70.46 & 49.09 & 67.64 & 67.41 & 63.06 \\
BERT-Base (FP) & \textbf{67.85} & \textbf{70.54} & 50.11 & \textbf{69.11} & 53.83 & 62.29 \\
BERT-Base (CP) & 59.68 & \textbf{70.54} & \textbf{55.06} & 67.42 & \textbf{69.49} & \textbf{64.44} \\
\hline
BERT-Large (VP) & 56.52 & \textbf{71.76} & 42.36 & 77.25 & 66.77 & 62.93 \\
BERT-Large (FP) & 66.17 & 64.70 & 50.37 & 65.44 & 52.24 & 59.78 \\
BERT-Large (CP) & \textbf{82.25} & 70.02 & \textbf{53.55} & \textbf{77.67} & \textbf{71.88} & \textbf{71.07} \\
\hline
RoBERTa-Base (VP) & 54.48 & 61.80 & 49.99 & 61.59 & 59.11 & 57.39 \\
RoBERTa-Base (FP) & \textbf{64.85} & \textbf{72.38} & 35.85 & \textbf{63.01} & 51.11 & 57.44 \\
RoBERTa-Base (CP) & 63.09 & 64.54 & \textbf{54.56} & 61.81 & \textbf{62.78} & \textbf{61.36} \\
\hline
RoBERTa-Large (VP) & 42.16 & 71.44 & 43.28 & \textbf{80.63} & 59.27 & 59.36 \\
RoBERTa-Large (FP) & 70.51 & 71.94 & 42.26 & 73.70 & 62.94 & 64.27 \\
RoBERTa-Large (CP) & \textbf{89.00} & \textbf{74.02} & \textbf{66.09} & 79.87 & \textbf{65.18} & \textbf{74.83} \\
\bottomrule
\end{tabular}
\end{center}
\vspace{-2mm}
\caption{Results of specificity testing with different prompts. The best results in each group are \textbf{bold}. VP: Vanilla Prompting, FP: Few-shot Prompting, CP: Cascade Prompting. \revise{*We do not rescore all suffixes for GPT-2 (CP).} 
}
\label{table:p_g}
\vspace{-5mm}
\end{table*}

\begin{table*}[h]
\begin{center}
\small
\setlength\tabcolsep{4pt} %
\begin{tabular}{l|r|r|r|r|r}
\toprule
& GPT-2 & BERT-Base & BERT-Large & RoBERTa-Base & RoBERTa-Large \\
\midrule 
\textit{Acc@$10$} w/ FP & + 10.62 & + 0.05 & + 2.74 & + 8.09 & + 16.45 \\
\textit{Acc@$10$} w/ CP & 0.00 & - 0.07 & - 4.28 & + 2.06 & + 0.77 \\
\bottomrule
\end{tabular}
\end{center}
\vspace{-2mm}
\caption{\revisenew{Change in correctness of the predictions compared to Vanilla Prompting $(\%)$ on fine-grained answers. \textit{w/ FP \& CP} means Few-shot \& Cascade Prompting is used to create prompts.}}
\label{table:correctness_fp_cp}
\vspace{-1mm}
\end{table*}

\begin{table*}[h]
\begin{center}
\small
\setlength\tabcolsep{4pt} %
\begin{tabular}{l|c|c|c|c|c}
\toprule
& GPT-2 & BERT-Base & BERT-Large & RoBERTa-Base & RoBERTa-Large \\
\midrule 
Naturalness w/ VP & 43.88 & 48.72 & 50.42 & 41.63 & 37.32 \\
\rowcolor{gray!20}
\cellcolor{white}Specificity w/ VP & 56.82 & 63.06 & 62.93 & 57.39 & 59.36 \\
\hline
Naturalness w/ FP & \uline{52.02} & 51.05 & 47.36 & \uline{49.11} & 49.96  \\
\rowcolor{gray!20}
\cellcolor{white}Specificity w/ FP & \textbf{64.96} & 62.29 & 59.78 & 57.44 & 64.27  \\
\hline  
Naturalness w/ CP & 43.88 & \uline{51.44} & \uline{56.54} & 45.81 & \uline{57.69}  \\
\rowcolor{gray!20}
\cellcolor{white}Specificity w/ CP & 56.82 & \textbf{64.44} & \textbf{71.07} & \textbf{61.36} & \textbf{74.83} \\
\bottomrule
\end{tabular}
\end{center}
\vspace{-1mm}
\caption{Average naturalness measured with $p_r (\%)$ with different prompts, with corresponding average specificity as reference. \textit{w/ VP} means Vanilla Prompting is used to create prompts. For each model, the best naturalness is \uline{underlined} and the best specificity is \textbf{bold}.
 }
\vspace{-4mm}
\label{table:naturalness}
\end{table*}

\subsection{Few-Shot Prompting}

We refer to using prompts in Table \ref{table:examples} to extract answers as \textbf{\textit{Vanilla Prompting}} \revise{(e.g., we let PLMs predict the masked token in ``John G. Bennett was born in [MASK].'')}. Vanilla Prompting cannot elicit specific answers since the designed prompts can not tell the models the preference regarding specificity; therefore, the models are not aware of whether a more specific answer is preferred. 

Based on the above analysis, we need to give the model some ``hints'' in terms of specificity, which can be achieved by providing some demonstrations. For instance, to predict where Toronto is located, if we provide some examples with coarse-grained answers using prompt ``\textcolor{brown}{Melbourne is located in Australia, Guangzhou is located in China,} \textcolor{violet}{Toronto is located in [MASK].}'', the model may know by analogy that we prefer a coarse-grained answer, which is Canada (a country).
In contrast, if we provide some fine-grained answers with ``\textcolor{brown}{Melbourne is located in Victoria, Guangzhou is located in Guangdong,} \textcolor{violet}{Toronto is located in [MASK].}'', the model may realize through analogy that we prefer a fine-grained answer here, which is Ontario (a province).

We refer to the method described above as \textbf{\textit{Few-shot Prompting}}, which supposes to bias the prediction to be more specific by providing some examples with fine-grained answers. The technique here is similar to the few-shot setting in GPT-3 \cite{NEURIPS2020_1457c0d6} and \cite{adolphs2021query}, where several demonstrations are given to the model as condition to help the model make the prediction.

\subsection{Cascade Prompting}

To make the answer more specific, we can also utilize the relationship between coarse-grained and fine-grained objects. For instance, in Table \ref{table:examples}, \textit{tracking ship} is a subclass of \textit{vessel}, while \textit{vessel} is also a subclass of \textit{vehicle}. To combine the three entities, we can write: \textit{Tracking ship is a subclass of vessel, which is a subclass of vehicle.} By masking the objects, we get prompt: ``\textcolor{violet}{Tracking ship is a subclass of [MASK],} \textcolor{brown}{which is a subclass of [MASK].}'' Intuitively, the first masked token will be more likely to be filled by \textit{vessel}, while the second masked token tends to be \textit{vehicle}.
Another example in Table \ref{table:examples} is to predict the birthplace, we can create prompt ``\textcolor{violet}{John G. Bennett was born in [MASK],} \textcolor{brown}{which is located in [MASK].}'' to bias the prediction of the first masked token to be more specific.

We refer to the above method as \textbf{\textit{Cascade Prompting}}, which aims to improve the specificity by adding ``\textit{which clauses}'' as constraints according to the relationship between coarse-grained and fine-grained answers. The ``which clauses'' here can be considered as suffixes and the prediction of the first masked token is returned as the answer.

\vspace{-1mm}
\section{Experiments}
\vspace{-1mm}

\label{sec:results}

In this section, we conduct experiments with the prompt-based methods proposed in Section \ref{sec:method}. 

\subsection{Experimental Setup}

We follow the setup in Section \ref{sec:setup}. For Few-shot Prompting, we set $K$, i.e., the number of demonstrations, as $10$. For Cascade Prompting, we apply the prompts in Table \ref{table:cascade_prompts}, which are constructed automatically based on the prompts for the transitive relations, e.g., ``... is located in [MASK].'' $\Rightarrow$ ``\textcolor{violet}{...,} \textcolor{brown}{which is located in [MASK].}''

\vspace{-1mm}
\subsection{Results}

Table \ref{table:p_g} summarizes the results of specificity testing with different prompting methods.
From the results, we observe that Cascade Prompting achieves the best performance in most cases. In addition, the performance improvement for BERT-Large and RoBERTa-Large with Cascade Prompting is quite significant. We think this is because the large models can understand which clauses better than the base models. 

We also observe that Few-shot Prompting does not always improve the specificity for bidirectional language models.
However, for GPT-2, which is a unidirectional language model, Few-shot Prompting achieves a significant performance improvement, while the results of Cascade Prompting are the same as those of Vanilla Prompting.

\revisenew{To observe the impact of the two methods on correctness, we report the change in correctness in Table \ref{table:correctness_fp_cp}. We observe that the correctness of Cascade Prompting is close to that of Vanilla Prompting, while the correctness of Few-shot Prompting improves significantly. This is because Cascade Prompting is in a zero-shot setting, while in Few-shot Prompting, demonstrations can provide some supervision to help the model make predictions.}

We also measure \textit{naturalness} of different models with different prompting methods. From Table \ref{table:naturalness}, we find that, for each model, the best prompting method is usually associated with the highest naturalness: 
Cascade Prompting improves the naturalness for bidirectional language models significantly, which corresponds to better performance on specificity; while for GPT-2, the naturalness using Few-shot Prompting is the highest, corresponding to the highest specificity.

\section{Discussion}

\label{sec:discussion}

{\flushleft \textbf{Specificity Testing in More General Scenarios}:} In this work, we test the specificity of PLMs on several relations with manually crafted prompts, with test data created automatically based on the property of transitive relations. For future work, we may test the specificity in more general scenarios. 
For instance, for numerical knowledge \cite{lin2020birds}, we can test how specifically PLMs describe the numbers, e.g., \textit{Obama was born in \uline{1961}} vs \textit{Obama was born in \uline{1960s}}, \textit{A car has \uline{four} wheels} vs \textit{A car has \uline{several} wheels}. In addition, we may test on multi-token answers \cite{jiang2020x}, and measure the specificity of sentences generated by PLMs \cite{louis2011automatic,ko2019domain,adiwardana2020towards,thoppilan2022lamda}, e.g., \textit{This is a very good paper. I really like it.} vs 
\textit{This paper conducts a very novel and interesting study, which provides a new insight for future work on language models.}

{\flushleft \textbf{Further Improvement of Specificity}:} In this paper, we propose \textit{Few-shot Prompting} and \textit{Cascade Prompting} to improve the specificity of PLMs without any additional training.
Future work may improve the specificity by including prompt-based fine-tuning \cite{shin2020eliciting,gao2021making}.
\revisenew{The observation also encourages future work to take into account the specificity, e.g., adding constraints regarding specificity, in the pre-training process.}
\revise{It is also interesting to design methods to control the degree of specificity for different usage scenarios \cite{huang2021definition}.}

\section{Conclusion}
\label{sec:conclusion}

In this paper, we build a benchmark to measure the specificity of predictions produced by pre-trained language models. 
To achieve this, we propose a novel approach to construct test data for specificity testing automatically.
\revisenew{From our evaluations, we show that existing PLMs have only a slight preference for more specific answers.}
We also propose two prompt-based methods, i.e., Few-shot Prompting and Cascade Prompting, to improve the specificity of the predictions. Extensive experiments and in-depth analyses demonstrate the effectiveness of the proposed methods.
We hope this work can encourage future research in this direction and give some insights to improve downstream tasks such as question answering, information extraction, and text generation: 1)~to make the answers, the extracted information, or the generated sentences more specific; 2)~to control the degree of specificity for different usage scenarios. 

\section*{Limitations}

This work presents some limitations. Firstly, our focus is confined to evaluating the specificity of predictions made by pre-trained language models for entity relations. As noted in Section~\ref{sec:discussion}, specificity can potentially be tested in a broader range of scenarios. Despite this restriction, we consider this work as an initial attempt to highlight the concept of language model specificity. We believe it will stimulate further research into this crucial, yet under-explored, area.

A second limitation is the scale of the models evaluated in this work. Given the swift evolution of large language models concurrent with the drafting of this paper, the models we examined are comparatively small.
As pointed out in the work of \citet{zheng2023does}, large language models may fail to answer a problem at the appropriate level of specificity.
We thus encourage future investigations to delve into the specificity of these rapidly evolving, larger language models.

\section*{Acknowledgements}

We thank the reviewers for their constructive feedback.
This material is based upon work supported by the National Science Foundation IIS 16-19302 and IIS 16-33755, Zhejiang University ZJU Research 083650, IBM-Illinois Center for Cognitive Computing Systems Research (C3SR) and IBM-Illinois Discovery Accelerator Institute (IIDAI), gift grants from eBay and Microsoft Azure, UIUC OVCR CCIL Planning Grant 434S34, UIUC CSBS Small Grant 434C8U, and UIUC New Frontiers Initiative. Any opinions, findings, and conclusions or recommendations expressed in this publication are those of the author(s) and do not necessarily reflect the views of the funding agencies.

\bibliography{custom}

\begin{thebibliography}{28}
\expandafter\ifx\csname natexlab\endcsname\relax\def\natexlab#1{#1}\fi

\bibitem[{Adiwardana et~al.(2020)Adiwardana, Luong, So, Hall, Fiedel,
  Thoppilan, Yang, Kulshreshtha, Nemade, Lu et~al.}]{adiwardana2020towards}
Daniel Adiwardana, Minh-Thang Luong, David~R So, Jamie Hall, Noah Fiedel, Romal
  Thoppilan, Zi~Yang, Apoorv Kulshreshtha, Gaurav Nemade, Yifeng Lu, et~al.
  2020.
\newblock Towards a human-like open-domain chatbot.
\newblock \emph{ArXiv preprint}, abs/2001.09977.

\bibitem[{Adolphs et~al.(2021)Adolphs, Dhuliawala, and
  Hofmann}]{adolphs2021query}
Leonard Adolphs, Shehzaad Dhuliawala, and Thomas Hofmann. 2021.
\newblock How to query language models?
\newblock \emph{arXiv preprint arXiv:2108.01928}.

\bibitem[{Bouraoui et~al.(2020)Bouraoui, Camacho-Collados, and
  Schockaert}]{bouraoui2020inducing}
Zied Bouraoui, Jose Camacho-Collados, and Steven Schockaert. 2020.
\newblock Inducing relational knowledge from bert.
\newblock In \emph{Proceedings of the AAAI Conference on Artificial
  Intelligence}, volume~34, pages 7456--7463.

\bibitem[{Brown et~al.(2020)Brown, Mann, Ryder, Subbiah, Kaplan, Dhariwal,
  Neelakantan, Shyam, Sastry, Askell, Agarwal, Herbert-Voss, Krueger, Henighan,
  Child, Ramesh, Ziegler, Wu, Winter, Hesse, Chen, Sigler, Litwin, Gray, Chess,
  Clark, Berner, McCandlish, Radford, Sutskever, and
  Amodei}]{NEURIPS2020_1457c0d6}
Tom Brown, Benjamin Mann, Nick Ryder, Melanie Subbiah, Jared~D Kaplan, Prafulla
  Dhariwal, Arvind Neelakantan, Pranav Shyam, Girish Sastry, Amanda Askell,
  Sandhini Agarwal, Ariel Herbert-Voss, Gretchen Krueger, Tom Henighan, Rewon
  Child, Aditya Ramesh, Daniel Ziegler, Jeffrey Wu, Clemens Winter, Chris
  Hesse, Mark Chen, Eric Sigler, Mateusz Litwin, Scott Gray, Benjamin Chess,
  Jack Clark, Christopher Berner, Sam McCandlish, Alec Radford, Ilya Sutskever,
  and Dario Amodei. 2020.
\newblock Language models are few-shot learners.
\newblock In \emph{Advances in Neural Information Processing Systems},
  volume~33, pages 1877--1901. Curran Associates, Inc.

\bibitem[{Devlin et~al.(2019)Devlin, Chang, Lee, and
  Toutanova}]{devlin2019bert}
Jacob Devlin, Ming-Wei Chang, Kenton Lee, and Kristina Toutanova. 2019.
\newblock Bert: Pre-training of deep bidirectional transformers for language
  understanding.
\newblock In \emph{Proceedings of the 2019 Conference of the North American
  Chapter of the Association for Computational Linguistics: Human Language
  Technologies, Volume 1 (Long and Short Papers)}, pages 4171--4186.

\bibitem[{Gao et~al.(2021)Gao, Fisch, and Chen}]{gao2021making}
Tianyu Gao, Adam Fisch, and Danqi Chen. 2021.
\newblock Making pre-trained language models better few-shot learners.
\newblock In \emph{Association for Computational Linguistics (ACL)}.

\bibitem[{Huang et~al.(2021)Huang, Kajiwara, and Arase}]{huang2021definition}
Han Huang, Tomoyuki Kajiwara, and Yuki Arase. 2021.
\newblock Definition modelling for appropriate specificity.
\newblock In \emph{Proceedings of the 2021 Conference on Empirical Methods in
  Natural Language Processing}, pages 2499--2509.

\bibitem[{Jiang et~al.(2020{\natexlab{a}})Jiang, Anastasopoulos, Araki, Ding,
  and Neubig}]{jiang2020x}
Zhengbao Jiang, Antonios Anastasopoulos, Jun Araki, Haibo Ding, and Graham
  Neubig. 2020{\natexlab{a}}.
\newblock X-factr: Multilingual factual knowledge retrieval from pretrained
  language models.
\newblock In \emph{Proceedings of the 2020 Conference on Empirical Methods in
  Natural Language Processing (EMNLP)}, pages 5943--5959.

\bibitem[{Jiang et~al.(2020{\natexlab{b}})Jiang, Xu, Araki, and
  Neubig}]{jiang2020can}
Zhengbao Jiang, Frank~F Xu, Jun Araki, and Graham Neubig. 2020{\natexlab{b}}.
\newblock How can we know what language models know?
\newblock \emph{Transactions of the Association for Computational Linguistics},
  8:423--438.

\bibitem[{Khashabi et~al.(2020)Khashabi, Min, Khot, Sabharwal, Tafjord, Clark,
  and Hajishirzi}]{khashabi2020unifiedqa}
Daniel Khashabi, Sewon Min, Tushar Khot, Ashish Sabharwal, Oyvind Tafjord,
  Peter Clark, and Hannaneh Hajishirzi. 2020.
\newblock Unifiedqa: Crossing format boundaries with a single qa system.
\newblock In \emph{Proceedings of the 2020 Conference on Empirical Methods in
  Natural Language Processing: Findings}, pages 1896--1907.

\bibitem[{Ko et~al.(2019)Ko, Durrett, and Li}]{ko2019domain}
Wei-Jen Ko, Greg Durrett, and Junyi~Jessy Li. 2019.
\newblock Domain agnostic real-valued specificity prediction.
\newblock In \emph{Proceedings of the AAAI Conference on Artificial
  Intelligence}, volume~33, pages 6610--6617.

\bibitem[{Lin et~al.(2020)Lin, Lee, Khanna, and Ren}]{lin2020birds}
Bill~Yuchen Lin, Seyeon Lee, Rahul Khanna, and Xiang Ren. 2020.
\newblock Birds have four legs?! numersense: Probing numerical commonsense
  knowledge of pre-trained language models.
\newblock In \emph{Proceedings of the 2020 Conference on Empirical Methods in
  Natural Language Processing (EMNLP)}, pages 6862--6868.

\bibitem[{Liu et~al.(2021{\natexlab{a}})Liu, Yuan, Fu, Jiang, Hayashi, and
  Neubig}]{liu2021pre}
Pengfei Liu, Weizhe Yuan, Jinlan Fu, Zhengbao Jiang, Hiroaki Hayashi, and
  Graham Neubig. 2021{\natexlab{a}}.
\newblock Pre-train, prompt, and predict: A systematic survey of prompting
  methods in natural language processing.
\newblock \emph{arXiv preprint arXiv:2107.13586}.

\bibitem[{Liu et~al.(2021{\natexlab{b}})Liu, Zheng, Du, Ding, Qian, Yang, and
  Tang}]{liu2021gpt}
Xiao Liu, Yanan Zheng, Zhengxiao Du, Ming Ding, Yujie Qian, Zhilin Yang, and
  Jie Tang. 2021{\natexlab{b}}.
\newblock Gpt understands, too.
\newblock \emph{arXiv preprint arXiv:2103.10385}.

\bibitem[{Liu et~al.(2019)Liu, Ott, Goyal, Du, Joshi, Chen, Levy, Lewis,
  Zettlemoyer, and Stoyanov}]{liu2019roberta}
Yinhan Liu, Myle Ott, Naman Goyal, Jingfei Du, Mandar Joshi, Danqi Chen, Omer
  Levy, Mike Lewis, Luke Zettlemoyer, and Veselin Stoyanov. 2019.
\newblock Roberta: A robustly optimized bert pretraining approach.
\newblock \emph{arXiv preprint arXiv:1907.11692}.

\bibitem[{Louis and Nenkova(2011)}]{louis2011automatic}
Annie Louis and Ani Nenkova. 2011.
\newblock Automatic identification of general and specific sentences by
  leveraging discourse annotations.
\newblock In \emph{Proceedings of 5th international joint conference on natural
  language processing}, pages 605--613.

\bibitem[{Marvin and Linzen(2018)}]{marvin2018targeted}
Rebecca Marvin and Tal Linzen. 2018.
\newblock Targeted syntactic evaluation of language models.
\newblock In \emph{Proceedings of the 2018 Conference on Empirical Methods in
  Natural Language Processing}, pages 1192--1202.

\bibitem[{Petroni et~al.(2019)Petroni, Rockt{\"a}schel, Riedel, Lewis, Bakhtin,
  Wu, and Miller}]{petroni2019language}
Fabio Petroni, Tim Rockt{\"a}schel, Sebastian Riedel, Patrick Lewis, Anton
  Bakhtin, Yuxiang Wu, and Alexander Miller. 2019.
\newblock Language models as knowledge bases?
\newblock In \emph{Proceedings of the 2019 Conference on Empirical Methods in
  Natural Language Processing and the 9th International Joint Conference on
  Natural Language Processing (EMNLP-IJCNLP)}, pages 2463--2473.

\bibitem[{Radford et~al.(2019)Radford, Wu, Child, Luan, Amodei, and
  Sutskever}]{radford2019language}
Alec Radford, Jeffrey Wu, Rewon Child, David Luan, Dario Amodei, and Ilya
  Sutskever. 2019.
\newblock Language models are unsupervised multitask learners.
\newblock \emph{OpenAI blog}, 1(8):9.

\bibitem[{Roberts et~al.(2020)Roberts, Raffel, and Shazeer}]{roberts2020much}
Adam Roberts, Colin Raffel, and Noam Shazeer. 2020.
\newblock How much knowledge can you pack into the parameters of a language
  model?
\newblock In \emph{Proceedings of the 2020 Conference on Empirical Methods in
  Natural Language Processing (EMNLP)}, pages 5418--5426.

\bibitem[{Shin et~al.(2020)Shin, Razeghi, Logan~IV, Wallace, and
  Singh}]{shin2020eliciting}
Taylor Shin, Yasaman Razeghi, Robert~L Logan~IV, Eric Wallace, and Sameer
  Singh. 2020.
\newblock Eliciting knowledge from language models using automatically
  generated prompts.
\newblock In \emph{Proceedings of the 2020 Conference on Empirical Methods in
  Natural Language Processing (EMNLP)}, pages 4222--4235.

\bibitem[{Thoppilan et~al.(2022)Thoppilan, De~Freitas, Hall, Shazeer,
  Kulshreshtha, Cheng, Jin, Bos, Baker, Du et~al.}]{thoppilan2022lamda}
Romal Thoppilan, Daniel De~Freitas, Jamie Hall, Noam Shazeer, Apoorv
  Kulshreshtha, Heng-Tze Cheng, Alicia Jin, Taylor Bos, Leslie Baker, Yu~Du,
  et~al. 2022.
\newblock Lamda: Language models for dialog applications.
\newblock \emph{arXiv preprint arXiv:2201.08239}.

\bibitem[{Vrande{\v{c}}i{\'c} and Kr{\"o}tzsch(2014)}]{vrandevcic2014wikidata}
Denny Vrande{\v{c}}i{\'c} and Markus Kr{\"o}tzsch. 2014.
\newblock Wikidata: a free collaborative knowledgebase.
\newblock \emph{Communications of the ACM}, 57(10):78--85.

\bibitem[{Wang et~al.(2020)Wang, Liu, and Song}]{wang2020language}
Chenguang Wang, Xiao Liu, and Dawn Song. 2020.
\newblock Language models are open knowledge graphs.
\newblock \emph{arXiv preprint arXiv:2010.11967}.

\bibitem[{Wang et~al.(2021)Wang, Thompson, and Iyyer}]{wang2021phrase}
Shufan Wang, Laure Thompson, and Mohit Iyyer. 2021.
\newblock Phrase-bert: Improved phrase embeddings from bert with an application
  to corpus exploration.

\bibitem[{Welleck et~al.(2019)Welleck, Brantley, Iii, and Cho}]{welleck2019non}
Sean Welleck, Kiant{\'e} Brantley, Hal~Daum{\'e} Iii, and Kyunghyun Cho. 2019.
\newblock Non-monotonic sequential text generation.
\newblock In \emph{International Conference on Machine Learning}, pages
  6716--6726. PMLR.

\bibitem[{Yu and Ettinger(2020)}]{yu2020assessing}
Lang Yu and Allyson Ettinger. 2020.
\newblock Assessing phrasal representation and composition in transformers.
\newblock In \emph{Proceedings of the 2020 Conference on Empirical Methods in
  Natural Language Processing (EMNLP)}, pages 4896--4907.

\bibitem[{Zheng et~al.(2023)Zheng, Huang, and Chang}]{zheng2023does}
Shen Zheng, Jie Huang, and Kevin Chen-Chuan Chang. 2023.
\newblock Why does chatgpt fall short in providing truthful answers?
\newblock \emph{ArXiv preprint}, abs/2304.10513.

\end{thebibliography}
\bibliographystyle{acl_natbib}

\end{document}